\documentclass{article}

     \PassOptionsToPackage{numbers, compress}{natbib}
 \usepackage[final]{tackling_climate_workshop_style}

\usepackage[utf8]{inputenc} % allow utf-8 input
\usepackage[T1]{fontenc}    % use 8-bit T1 fonts
\usepackage{hyperref}       % hyperlinks
\usepackage{url}            % simple URL typesetting
\usepackage{booktabs}       % professional-quality tables
\usepackage{amsfonts}       % blackboard math symbols
\usepackage{nicefrac}       % compact symbols for 1/2, etc.
\usepackage{microtype}      % microtypography
\usepackage[pdftex]{graphicx}
\usepackage{xcolor} % colors in text only
\usepackage{caption}
\usepackage{subcaption}
\usepackage{wrapfig}
\usepackage{comment}
\usepackage[symbol]{footmisc}

\title{Identification of medical devices using machine learning on distribution feeder data for informing power outage response}

\author{
  Paraskevi Kourtza* \\
  University of Edinburgh \\
  Edinburgh, UK \\
  \texttt{s1265437@ed.ac.uk} \\
  \And
  Maitreyee Marathe* \\
  University of Wisconsin-Madison \\
  Madison, USA \\
  \texttt{mmarathe@wisc.edu} \\
  \And 
  Anuj Shetty* \\
  Stanford University\\
  Stanford, USA \\
  \texttt{anuj42@stanford.edu} \\
  \And
  Diego Kiedanski \\
  Yale University \\
  New Haven, USA \\
  \texttt{diego.kiedanski@yale.edu} \\
}

\begin{document}

\maketitle

\footnotetext{* These authors contributed equally.}

\begin{abstract}

Power outages caused by extreme weather events due to climate change have doubled in the United States in the last two decades. Outages pose severe health risks to over 4.4 million individuals dependent on in-home medical devices. Data on the number of such individuals residing in a given area is limited. This study proposes a load disaggregation model to predict the number of medical devices behind an electric distribution feeder. This data can be used to inform planning and response. The proposed solution serves as a measure for climate change adaptation.

\end{abstract}

\section{Problem and motivation}

Over 4.4 million people in the U.S. rely on electricity-dependent in-home medical devices and services \citep{noauthor_hhs_nodate}. Extreme weather events that cause power outages can lead to increased mortality rates \citep{kishore_mortality_2018, issa_deaths_2018} in such individuals or to additional stress for hospitals, shelters, and emergency services \citep{higgs_power_2009, nakayama_effect_2014, greenwald_emergency_2004}. Unfortunately, there have been several disaster emergencies that corroborate this (see Appendix \ref{appendix a-extreme events} for examples). Energy resilience for home healthcare is a largely unexplored problem and needs proactive approaches to meet the needs of this vulnerable community \citep{marathe_energy_homehealthcare}.

Climate change is driving more intense and frequent extreme weather events, which put pressure on an aging power grid infrastructure. Power outages tied to extreme weather have doubled across the U.S. in the last two decades, and the frequency and length of outages are at their highest \citep{brown_storms_2022}. Also, the lack of comprehensive data on medically fragile individuals hinders effective planning and response.

We propose performing load disaggregation on electric distribution feeder data to identify the number of in-home medical devices downstream of a feeder. Our solution does not depend on precise location or number of individuals, therefore respecting personal privacy. We also propose creating a dataset of power data measurements for in-home medical devices. Our method can be used to estimate the number of medically fragile individuals in an area and improve preparedness for power outages.

%It is necessary to have a comprehensive dataset of the number of such medically fragile individuals and their locations for effective resource planning and response to power outages and extreme weather events. We propose load disaggregation on electric distribution feeder data to identify the number of in-home medical devices present downstream of the feeder. This will not reveal the precise location or number of individuals, protecting their privacy, but will offer an estimate of the number of medical devices present in the area served by the feeder. This data can be useful during extreme weather event induced power outages and other emergencies, and is an approach for climate change adaptation.

\section{Related approaches and datasets}

Traditional methods to find the number and locations of such individuals rely on private patient data, either on lists by healthcare providers or electric utility medical baseline programs (e.g., from PG\&E \citep{noauthor_medical_nodate}), where customers can self-register if they use in-home medical devices.  However, this is relevant only for countries with centralized healthcare and it can lead to incomplete and inaccurate data, as shown by a 2018 survey on New York City residents \citep{dominianni_power_2018}.

The HHS emPOWER map \citep{noauthor_hhs_nodate} has improved on utility baseline programs by using Medicare insurance data to estimate the number of people who rely on in-home medical devices at zip code level. The map covers Medicare beneficiaries but it is estimated that millions more rely on electricity-dependent medical technology in the U.S., including over 180,000 children \citep{shapiro_home_2019}. The proposed study aims to build upon and expand the emPOWER map.

We found only one method in the literature that uses energy consumption data to address a facet of this problem. \citet{bean_keeping_2020} assume prior knowledge of the houses with in-home medical devices, and use smart meter voltage data to identify their supply phase. Our approach is based on energy disaggregation at the feeder level. Load disaggregation has been reviewed in \citep{angelis_nilm_2022, kelly_neural_2015} and most related work addresses disaggregation at the household level. \citet{ledva_real-time_2018} perform feeder-level disaggregation, using real-time feeder values to separate the demand of air conditioners and of other loads. Such non-intrusive load monitoring (NILM) techniques have also recently grown in popularity across different use cases, such as estimating solar PV generation for a given feeder \citep{vrettos_pv_2019}. However, these methods rely on real-time feedback for forecasting, which does not apply to our problem. Further discussion on related approaches can be found in Appendix \ref{appendix:related-approaches}

%\citet{bean_keeping_2020} assume prior knowledge of which houses have in-home medical devices and use smart meter voltage data at 15-minute resolution. They use an unsupervised learning algorithm to cluster houses with correlated voltage time series and identify the phase houses with medical devices are on. With this information, network operators can minimize the likelihood of loss of power to in-home medical device users

\section{Proposed methodology}
\begin{wrapfigure}{r}{0.45\textwidth}
  \begin{center}
  \vspace{-8mm}
  \includegraphics[width = 0.4\textwidth]{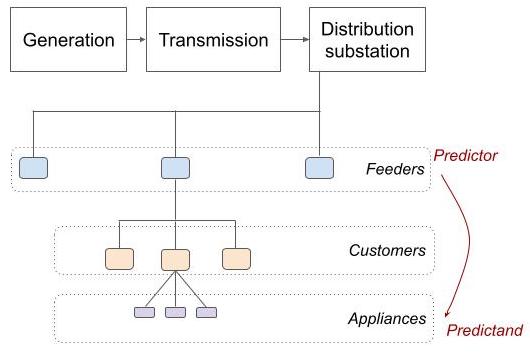}
  \caption{Distribution system}
  \vspace{-3mm}
  \label{fig:distribution_system}
  \end{center}
\end{wrapfigure}

The input for the proposed model is power data from an electric distribution system feeder. As shown in Figure  \ref{fig:distribution_system}, a distribution substation has many feeders, and each feeder has multiple customers downstream. The predictor of the model is power data measured at the feeder, and the predictand is the number of medical devices present downstream of that feeder.

We propose partnering with 10-20 households using in-home medical devices and gaining consent to monitor their device usage. This range is large enough for a proof of concept, while being small enough to be feasible to locate within a feeder and onboard for our data collection. However, this may change at the time of implementation. Monitoring would be through smart meters at the medical device level, with a sampling rate of the order of 1Hz. These measurements would be used mainly to record whether each device was on or off for each timestamp, given some threshold. We will get a time series of the number of such devices running at a time to be used as ground truth data for training, and this would not be required at the time of model inference. The set of in-home medical devices is vast \citep{unitedhealthcare}, and we propose starting with a single device as a proof of concept. We propose choosing the in-home ventilator \citep{vyairemedical} as the device of interest, since it is used for pediatric patients as well as older adults. \citep{King921}.

\begin{figure}[htbp]
  \centering
  \vspace{-5mm}
  \includegraphics[width=0.7\linewidth]{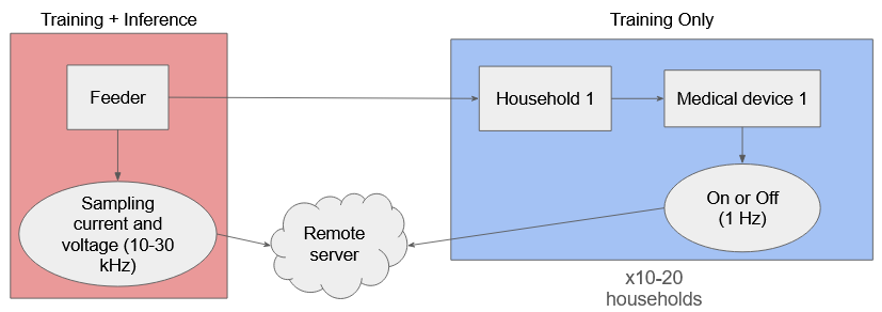}
  %\fbox{\rule[-.5cm]{0cm}{4cm} \rule[-.5cm]{4cm}{0cm}}
  \caption{Dataset generation}
  \label{fig:dataset_generation}
\end{figure}

As shown in Figure \ref{fig:dataset_generation}, for the training input, we would also need to partner with the distribution company to sample the feeder current with an Analog to Digital card at a sampling rate of 10-20 kHz. We assume we can identify and monitor all medical devices of the selected type served by the feeder. This data collection would be carried out over the course of 1 week. This duration is chosen based on other standard high-frequency datasets like BLUED \citep{anderson2012blued}.

% Our model has a multivariate input consisting of 1) the power time series and 2) high-frequency features to help distinguish the medical device type.
Let $i_t, v_t$ denote the instantaneous current and voltage measurement\footnote{Either single or multiphase.}. To transform a time-series problem into a tabular format compatible with traditional neural architectures, we will consider a rolling window of size $W$ (e.g., 5 seconds): $I^W_t = [i_{t-W+1}, \dots, i_t]$ and $V^W_t = [v_{t-W+1}, \dots, v_t]$. 
The final dataset $X$ will be a collection of vectors $x_t$ where each $x_t$ is itself a collection of features obtained from $I^W_t$ and $V^W_t$, i.e.,

\begin{equation}
x_t = (f^1(I^W_t, V^W_t), \dots,  f^L(I^W_t, V^W_t)))
\label{eq:input}
\end{equation}

with $f^j\colon \mathbb{R}^{W} \times \mathbb{R}^{W} \to \mathbb{R}$. 
To obtain the feature functions $f^j$, we will test the medical device in the lab in all its different operational modes. We would then pick the most essential features to distinguish this device's signature from other device types found in the PLAID dataset \citep{medico_voltage_2020}, as used in \citep{marchesoni-acland_end--end_2020}. These high-frequency features include the form factor, the phase shift between voltage and current, etc., similar to \citep{marchesoni-acland_end--end_2020}.

% These would be calculated over rolling time windows and then fed into the convolution layers of the model, 

% \begin{figure}[h!]
%     \centering
%     \includegraphics[width=0.3\linewidth]{figures/VI_trajectory_example_for_different_devices1.png}
%     \hspace{0.1cm}
%     \includegraphics[width=0.3\linewidth]{figures/VI_trajectory_example_for_different_devices2.png}
%     \hspace{0.1cm}
%     \includegraphics[width=0.3\linewidth]{figures/VI_trajectory_example_for_different_devices3.png}
%     \caption{VI trajectory example for different devices \cite{marchesoni-acland_end--end_2020}}
%     \label{fig:VI_trajectory}
% \end{figure}

% \begin{equation}
% % x_t = (p_{0,0}, \dots, p_{0, M}, \dots, p_{M, N}, ff, \dots), \ , y_t = (?)
% % x_t = (p_t, f_{1,t}, \dots, f_{N,t} )
% X = (x_0, \dots, x_{T-1}), \ , x_t = (f^1(I_t, V_t), \dots,  f^L(I_t, V_t))), \ , I_t = [i_t, \dots, i_{t-w}], \ V_t = [v_t, \dots, v_{t-w}]
% \label{eq:input}
% \end{equation}
% \begin{equation}
% % f_{i,t} = f(i_{t-w/2}, v_{t-w/2}, \dots, i_{t+w/2}, v_{t+w/2})
% \label{eq:feature}
% \end{equation}
% \begin{equation}
% y_t = \sum\limits_{j=1}^{N}d_{j,t}
% \label{eq:output}
% \end{equation}

% The input to the model $x_t$ is given by equation \ref{eq:input} where {\color{blue}$p_t$ is ?} and $f_{i,t}$ is a high frequency feature calculated over a window $w$ around time $t$ as given in equation \ref{eq:feature}. 

For every vector $x_t$, its corresponding target $y_t$ will correspond to the number of medical devices running within the time window $[t-W+1, t]$.

\begin{figure}[h!]
  \centering
  \vspace{-5mm}
  \includegraphics[width=0.6\linewidth]{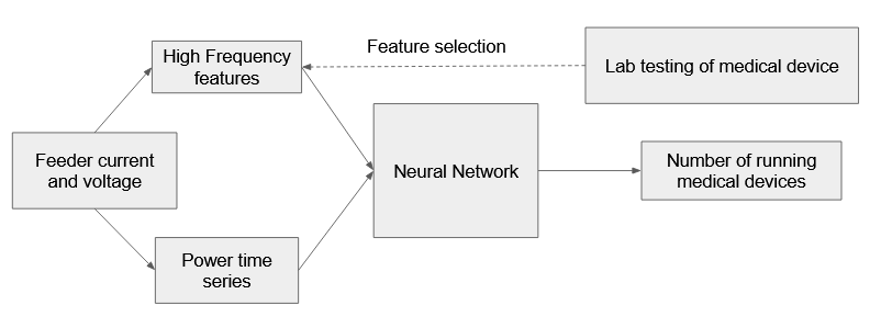}
  %\fbox{\rule[-.5cm]{0cm}{4cm} \rule[-.5cm]{4cm}{0cm}}
  \caption{Proposed model block diagram}
  \vspace{-5mm}
  \label{fig:model_block_diagram}
\end{figure}

% The output $y_t$ is given by equation \ref{eq:output}, where $N$ is the number of houses and $d_{j,t}$ is the number of medical devices running in house $j$ at time $t$. 
The proposed data pipeline and model is shown in Figure \ref{fig:model_block_diagram}, and is based on the neural network model used in \citep{marchesoni-acland_end--end_2020}. We use the Mean Absolute Error metric $
MAE = \frac{1}{N} \sum\limits_t | \hat{y_t} - y_t|$

\begin{comment}
\section{Discussion and deployment}
{\color{blue} Technical limitations (can be put into the Challenges and Limitations appendix):}\\
The model will be trained to predict the number of devices belonging to a set of common in-home medical devices. Therefore, if none of the devices an individual owns belong to this set, the individual will not be accounted for. This will limit the transferability of this model to a case where devices other than those considered need to be identified. The presented approach aims at extending the load disaggregation framework for feeder-level data. Although feeder level disaggregation has been implemented, for instance to identify air conditioning loads \citep{ledva_real-time_2018},  it remains to be seen if this can be successfully implemented for medical devices, particularly if the power signatures of other devices are similar to the target medical devices.
\end{comment}

\section{Impact and conclusions}

The contributions of this study would be (1) a load disaggregation model to predict the number of medical devices downstream of a feeder, (2) a quantitative approach to estimate the medically fragile population in an area while preserving location privacy of such individuals, (3) a dataset of high frequency power data measurements for in-home medical devices. The main challenges and limitations of this approach are discussed in Appendix \ref{appendix:challenges-limitations}.

The estimated data on medical devices in an area can be used by multiple entities to inform planning and response to extreme weather events and power outages, thereby functioning as a vehicle for climate change adaptation. It can be used by electric utilities to plan public safety power shut-offs \citep{noauthor_utility_nodate}, the capacity and locations of community charging stations in outage-prone areas, and for priority restoration after an outage. Home healthcare agencies can use this data to plan for supplies such as oxygen tanks. Emergency management services can use the data for public outreach and distribution of emergency kits before an extreme-weather event. The resulting data from the study can be used to complement the existing emPOWER dataset, which has already proven useful in adaptation to extreme weather events \citep{noauthor_story_2021, noauthor_story_2021-1, noauthor_story_2021-2, noauthor_story_2021-3}. 

There are multiple stakeholders in this space, namely medically fragile households, home healthcare agencies and hospitals, durable medical equipment providers, electric utilities, and emergency management services. Partnerships with these stakeholders are critical for the successful formulation and implementation of the proposed solution. Energy resilience for home healthcare is a largely unexplored problem, and the proposed model aims to address gaps at this intersection.

\section*{Acknowledgements}
The authors would like to thank the organizers of the Climate Change AI Summer School 2022 for facilitating the initial work for this proposal.

% Use unnumbered first-level headings for the acknowledgments. All acknowledgments
% More information about this disclosure can be found at: \url{https://neurips.cc/Conferences/2020/PaperInformation/FundingDisclosure}.

% Do {\bf not} include this section in the anonymized submission, only in the final paper. You can use the \texttt{ack} environment provided in the style file to automatically hide this section in the anonymized submission.

\medskip

\small

\bibliography{references}

\begin{thebibliography}{32}
\providecommand{\natexlab}[1]{#1}
\providecommand{\url}[1]{\texttt{#1}}
\expandafter\ifx\csname urlstyle\endcsname\relax
  \providecommand{\doi}[1]{doi: #1}\else
  \providecommand{\doi}{doi: \begingroup \urlstyle{rm}\Url}\fi

\bibitem[noa({\natexlab{a}})]{noauthor_hhs_nodate}
{HHS} {emPOWER} {Program}, {\natexlab{a}}.
\newblock URL \url{https://empowerprogram.hhs.gov/}.

\bibitem[Kishore et~al.(2018)Kishore, Marqués, Mahmud, V.~Kiang, Rodriguez,
  Fuller, Ebner, Sorensen, Racy, Lemery, Maas, and
  Leaning]{kishore_mortality_2018}
Nishant Kishore, Domingo Marqués, Ayesha Mahmud, Mathew V.~Kiang, Irmary
  Rodriguez, Arlan Fuller, Peggy Ebner, Cecilia Sorensen, Fabio Racy, Jay
  Lemery, Leslie Maas, and Jennifer Leaning.
\newblock Mortality in {Puerto} {Rico} after {Hurricane} {Maria}.
\newblock \emph{New England Journal of Medicine}, 379\penalty0 (17):\penalty0
  e30, October 2018.
\newblock ISSN 0028-4793, 1533-4406.
\newblock \doi{10.1056/NEJMc1810872}.
\newblock URL \url{http://www.nejm.org/doi/10.1056/NEJMc1810872}.

\bibitem[Issa(2018)]{issa_deaths_2018}
Anindita Issa.
\newblock Deaths {Related} to {Hurricane} {Irma} — {Florida}, {Georgia}, and
  {North} {Carolina}, {September} 4–{October} 10, 2017.
\newblock \emph{MMWR. Morbidity and Mortality Weekly Report}, 67, 2018.
\newblock ISSN 0149-21951545-861X.
\newblock \doi{10.15585/mmwr.mm6730a5}.
\newblock URL \url{https://www.cdc.gov/mmwr/volumes/67/wr/mm6730a5.htm}.

\bibitem[Higgs(2009)]{higgs_power_2009}
Robert Higgs.
\newblock Power outage potentially deadly for those on home life-support,
  January 2009.
\newblock URL
  \url{https://www.cleveland.com/health/2009/01/power_outage_potentially_deadl.html}.

\bibitem[Nakayama et~al.(2014)Nakayama, Tanaka, Uematsu, Kikuchi, Hino-Fukuyo,
  Morimoto, Sakamoto, Tsuchiya, and Kure]{nakayama_effect_2014}
Tojo Nakayama, Soichiro Tanaka, Mitsugu Uematsu, Atsuo Kikuchi, Naomi
  Hino-Fukuyo, Tetsuji Morimoto, Osamu Sakamoto, Shigeru Tsuchiya, and Shigeo
  Kure.
\newblock Effect of a blackout in pediatric patients with home medical devices
  during the 2011 eastern {Japan} earthquake.
\newblock \emph{Brain and Development}, 36\penalty0 (2):\penalty0 143--147,
  February 2014.
\newblock ISSN 0387-7604.
\newblock \doi{10.1016/j.braindev.2013.02.001}.
\newblock URL
  \url{https://www.sciencedirect.com/science/article/pii/S038776041300079X}.

\bibitem[Greenwald et~al.(2004)Greenwald, Rutherford, Green, and
  Giglio]{greenwald_emergency_2004}
Peter~W. Greenwald, Anne~F. Rutherford, Robert~A. Green, and James Giglio.
\newblock Emergency {Department} {Visits} for {Home} {Medical} {Device}
  {Failure} during the 2003 {North} {America} {Blackout}.
\newblock \emph{Academic Emergency Medicine}, 11\penalty0 (7):\penalty0
  786--789, July 2004.
\newblock ISSN 10696563, 15532712.
\newblock \doi{10.1197/j.aem.2003.12.032}.
\newblock URL \url{http://doi.wiley.com/10.1197/j.aem.2003.12.032}.

\bibitem[Marathe and Manur(2020)]{marathe_energy_homehealthcare}
Maitreyee Marathe and Ashray Manur.
\newblock Energy resilience for home healthcare, 2020.
\newblock URL \url{smplabs.wisc.edu/nsf-icorps}.

\bibitem[Brown et~al.(2022)Brown, Fassett, Whittle, McConnaughey, and
  Lo]{brown_storms_2022}
Matthew Brown, Camille Fassett, Patrick Whittle, Janet McConnaughey, and Jason
  Lo.
\newblock Storms batter aging power grid as climate disasters spread, April
  2022.
\newblock URL
  \url{https://apnews.com/article/wildfires-storms-science-business-health-7a0fb8c998c1d56759989dda62292379}.

\bibitem[noa({\natexlab{b}})]{noauthor_medical_nodate}
Medical {Baseline} {Program}, {\natexlab{b}}.
\newblock URL
  \url{https://www.pge.com/en_US/residential/save-energy-money/help-paying-your-bill/longer-term-assistance/medical-condition-related/medical-baseline-allowance/medical-baseline-allowance.page}.

\bibitem[Dominianni et~al.(2018)Dominianni, Ahmed, Johnson, Blum, Ito, and
  Lane]{dominianni_power_2018}
Christine Dominianni, Munerah Ahmed, Sarah Johnson, Micheline Blum, Kazuhiko
  Ito, and Kathryn Lane.
\newblock Power {Outage} {Preparedness} and {Concern} among {Vulnerable} {New}
  {York} {City} {Residents}.
\newblock \emph{Journal of Urban Health}, 95\penalty0 (5):\penalty0 716--726,
  October 2018.
\newblock ISSN 1099-3460, 1468-2869.
\newblock \doi{10.1007/s11524-018-0296-9}.
\newblock URL \url{http://link.springer.com/10.1007/s11524-018-0296-9}.

\bibitem[Shapiro and Mango(2019)]{shapiro_home_2019}
Annie Shapiro and Marriele Mango.
\newblock Home {Health} {Care} in the {Dark}: {Why} {Climate}, {Wildfires} and
  {Other} {Risks} {Call} for {New} {Resilient} {Energy} {Storage} {Solutions}
  to {Protect} {Medically} {Vulnerable} {Households} {From} {Power} {Outages}.
\newblock Technical report, Clean Energy Group, Meridian Institute, April 2019.
\newblock URL
  \url{https://www.cleanegroup.org/ceg-resources/resource/battery-storage-home-healthcare/}.

\bibitem[Bean et~al.(2020)Bean, Snow, Glencross, Viller, and
  Horrocks]{bean_keeping_2020}
Richard Bean, Stephen Snow, Mashhuda Glencross, Stephen Viller, and Neil
  Horrocks.
\newblock Keeping the power on to home medical devices.
\newblock \emph{PLoS ONE}, 15\penalty0 (7):\penalty0 e0235068, July 2020.
\newblock ISSN 1932-6203.
\newblock \doi{10.1371/journal.pone.0235068}.
\newblock URL \url{https://www.ncbi.nlm.nih.gov/pmc/articles/PMC7347141/}.

\bibitem[Angelis et~al.(2022)Angelis, Timplalexis, Krinidis, Ioannidis, and
  Tzovaras]{angelis_nilm_2022}
Georgios-Fotios Angelis, Christos Timplalexis, Stelios Krinidis, Dimosthenis
  Ioannidis, and Dimitrios Tzovaras.
\newblock {NILM} applications: {Literature} review of learning approaches,
  recent developments and challenges.
\newblock \emph{Energy and Buildings}, 261:\penalty0 111951, April 2022.
\newblock ISSN 03787788.
\newblock \doi{10.1016/j.enbuild.2022.111951}.
\newblock URL
  \url{https://linkinghub.elsevier.com/retrieve/pii/S0378778822001220}.

\bibitem[Kelly and Knottenbelt(2015{\natexlab{a}})]{kelly_neural_2015}
Jack Kelly and William Knottenbelt.
\newblock Neural {NILM}: {Deep} {Neural} {Networks} {Applied} to {Energy}
  {Disaggregation}.
\newblock In \emph{Proceedings of the 2nd {ACM} {International} {Conference} on
  {Embedded} {Systems} for {Energy}-{Efficient} {Built} {Environments}}, pages
  55--64, Seoul South Korea, November 2015{\natexlab{a}}. ACM.
\newblock ISBN 9781450339810.
\newblock \doi{10.1145/2821650.2821672}.
\newblock URL \url{https://dl.acm.org/doi/10.1145/2821650.2821672}.

\bibitem[Ledva et~al.(2018)Ledva, Balzano, and Mathieu]{ledva_real-time_2018}
Gregory~S. Ledva, Laura Balzano, and Johanna~L. Mathieu.
\newblock Real-{Time} {Energy} {Disaggregation} of a {Distribution} {Feeder}'s
  {Demand} {Using} {Online} {Learning}.
\newblock \emph{IEEE Transactions on Power Systems}, 33\penalty0 (5):\penalty0
  4730--4740, September 2018.
\newblock ISSN 0885-8950, 1558-0679.
\newblock \doi{10.1109/TPWRS.2018.2800535}.
\newblock URL \url{https://ieeexplore.ieee.org/document/8276574/}.

\bibitem[Vrettos(2019)]{vrettos_pv_2019}
E~Vrettos.
\newblock Estimating pv power from aggregate power measurements within the
  distribution grid.
\newblock \emph{Journal of Renewable and Sustainable Energy}, 1:\penalty0
  027307, 2019.
\newblock \doi{10.1063/1.5094161}.
\newblock URL \url{https://aip.scitation.org/doi/full/10.1063/1.5094161}.

\bibitem[uni(2022)]{unitedhealthcare}
United {H}ealthcare: Durable medical equipment reference list, 2022.
\newblock URL
  \url{https://www.uhcprovider.com/content/dam/provider/docs/public/policies/medadv-guidelines/d/durable-medical-equipment-dme-reference-list.pdf}.

\bibitem[vya()]{vyairemedical}
Vyaire {M}edical: {LTV} ventilator series.
\newblock URL
  \url{https://www.vyaire.com/sites/default/files/2020-08/vyr-gbl-2000203-ltv2-comparison-spec-sheet_2.0_final.pdf}.

\bibitem[King(2012)]{King921}
Angela~C King.
\newblock Long-term home mechanical ventilation in the united states.
\newblock \emph{Respiratory Care}, 57\penalty0 (6):\penalty0 921--932, 2012.
\newblock \doi{10.4187/respcare.01741}.

\bibitem[Anderson et~al.(2012)Anderson, Ocneanu, Benitez, Carlson, Rowe, and
  Berges]{anderson2012blued}
Kyle Anderson, Adrian Ocneanu, Diego Benitez, Derrick Carlson, Anthony Rowe,
  and Mario Berges.
\newblock Blued: A fully labeled public dataset for event-based non-intrusive
  load monitoring research.
\newblock pages 1--5, 2012.

\bibitem[Medico et~al.(2020)Medico, De~Baets, Gao, Giri, Kara, Dhaene,
  Develder, Bergés, and Deschrijver]{medico_voltage_2020}
Roberto Medico, Leen De~Baets, Jingkun Gao, Suman Giri, Emre Kara, Tom Dhaene,
  Chris Develder, Mario Bergés, and Dirk Deschrijver.
\newblock A voltage and current measurement dataset for plug load appliance
  identification in households.
\newblock \emph{Scientific Data}, 7\penalty0 (1):\penalty0 49, February 2020.
\newblock ISSN 2052-4463.
\newblock \doi{10.1038/s41597-020-0389-7}.

\bibitem[Marchesoni-Acland et~al.(2020)Marchesoni-Acland, Mariño, Masquil,
  Masaferro, and Fernández]{marchesoni-acland_end--end_2020}
Franco Marchesoni-Acland, Camilo Mariño, Elías Masquil, Pablo Masaferro, and
  Alicia Fernández.
\newblock End-to-end {NILM} {System} {Using} {High} {Frequency} {Data} and
  {Neural} {Networks}, April 2020.
\newblock URL \url{http://arxiv.org/abs/2004.13905}.
\newblock arXiv:2004.13905 [eess].

\bibitem[noa({\natexlab{c}})]{noauthor_utility_nodate}
Utility {Public} {Safety} {Power} {Shutoff} {Plans}, {\natexlab{c}}.
\newblock URL \url{https://www.cpuc.ca.gov/psps/}.

\bibitem[noa(2021{\natexlab{a}})]{noauthor_story_2021}
Story from the field - {Wildfires} in {Los} {Angeles} {County}, {California},
  September 2021{\natexlab{a}}.
\newblock URL \url{https://empowerprogram.hhs.gov/Wildfires-LA-County.pdf}.

\bibitem[noa(2021{\natexlab{b}})]{noauthor_story_2021-1}
Story from the field - {Earthquakes} in {Puerto} {Rico}, September
  2021{\natexlab{b}}.
\newblock URL \url{https://empowerprogram.hhs.gov/Earthquakes-Puerto-Rico.pdf}.

\bibitem[noa(2021{\natexlab{c}})]{noauthor_story_2021-2}
Story from the field - {Flooding} and windstorms in {New} {York}, September
  2021{\natexlab{c}}.
\newblock URL
  \url{https://empowerprogram.hhs.gov/Flooding-Windstorms-in-New-York.pdf}.

\bibitem[noa(2021{\natexlab{d}})]{noauthor_story_2021-3}
Story from the field - {Multiple} threats in {Arizona}, September
  2021{\natexlab{d}}.
\newblock URL
  \url{https://empowerprogram.hhs.gov/Multiple-Threats-in-Arizona.pdf}.

\bibitem[Pandey and Karypis(2019)]{pandey_structured_2019}
Shalini Pandey and George Karypis.
\newblock Structured {Dictionary} {Learning} for {Energy} {Disaggregation}.
\newblock In \emph{Proceedings of the {Tenth} {ACM} {International}
  {Conference} on {Future} {Energy} {Systems}}, pages 24--34, June 2019.
\newblock \doi{10.1145/3307772.3328301}.
\newblock URL \url{http://arxiv.org/abs/1907.06581}.
\newblock arXiv:1907.06581 [cs, eess].

\bibitem[Elhamifar and Sastry(2015)]{elhamifar_energy_2015}
Ehsan Elhamifar and Shankar Sastry.
\newblock Energy disaggregation via learning '{Powerlets}' and sparse coding.
\newblock In \emph{Proceedings of the {Twenty}-{Ninth} {AAAI} {Conference} on
  {Artificial} {Intelligence}}, {AAAI}'15, pages 629--635, Austin, Texas,
  January 2015. AAAI Press.
\newblock ISBN 9780262511292.

\bibitem[noa({\natexlab{d}})]{noauthor_dataport_nodate}
Dataport, {\natexlab{d}}.
\newblock URL \url{https://www.pecanstreet.org/dataport/}.

\bibitem[Kelly and Knottenbelt(2015{\natexlab{b}})]{kelly_uk-dale_2015}
Jack Kelly and William Knottenbelt.
\newblock The {UK}-{DALE} dataset, domestic appliance-level electricity demand
  and whole-house demand from five {UK} homes.
\newblock \emph{Scientific Data}, 2\penalty0 (1):\penalty0 150007, December
  2015{\natexlab{b}}.
\newblock ISSN 2052-4463.
\newblock \doi{10.1038/sdata.2015.7}.
\newblock URL \url{http://arxiv.org/abs/1404.0284}.
\newblock arXiv:1404.0284 [cs].

\bibitem[Bucci et~al.(2021)Bucci, Ciancetta, Fiorucci, Mari, and
  Fioravanti]{bucci_cnn_2021}
Giovanni Bucci, Fabrizio Ciancetta, Edoardo Fiorucci, Simone Mari, and Andrea
  Fioravanti.
\newblock Multi-state appliances identification through a nilm system based on
  convolutional neural network.
\newblock pages 1--6, 2021.
\newblock \doi{10.1109/I2MTC50364.2021.9460038}.

\end{thebibliography}

\appendix
\section{Examples of extreme weather events and their impact} \label{appendix a-extreme events}
One-third of 4654 additional deaths in Puerto Rico during the three months following Hurricane Maria in 2017 can be attributed to health complications due to outage-related problems, such as failure of in-home medical devices \citep{kishore_mortality_2018}. Over 15\% of the deaths after Hurricane Irma in 2017 were due to worsening pre-existing medical conditions because of power outages \citep{issa_deaths_2018}. After Hurricane Gustav, 20\% to 40\% of the 1400 people who came into medical shelters depended on medical equipment \citep{higgs_power_2009}. After the 2011 earthquakes in Japan, there was an influx of medical device-dependent individuals into hospitals \citep{nakayama_effect_2014}. In the 24 hours after the 2003 blackouts in New York, 22\% of hospital admits were people relying on in-home medical devices \citep{greenwald_emergency_2004}. 

\section{More on related approaches}
\label{appendix:related-approaches}
\citet{pandey_structured_2019} perform energy disaggregation at the household level. More specifically, they build upon Powerlet-based energy disaggregation (PED) \citep{elhamifar_energy_2015}, which “\textit{captures the different power consumption patterns of each appliance as representatives (used as dictionary atoms) and then estimates a combination of these representatives that best approximates the observed aggregated power consumption.}” To overcome the limitation of co-occurrence in PED, they use the information on operation modes of the different devices and distinguish between similar co-occurring devices. Devices are divided into subsets, forming a binary tree, until only one device is left. Decomposing the home energy consumption becomes a recursive task, starting from the root node. However, disaggregating every other device at the feeder level would be inefficient for our problem, as we only want to account for medical devices.

\citet{ledva_real-time_2018} perform feeder-level disaggregation, using real-time feeder measurements to separate the demand of air conditioners (AC) and the demand of other loads (OL). Their implementation is based on an online learning model, Dynamic Fixed Share (DFS), and models created from historic building-level and device-level data. The historical data have a sampling frequency of one minute and are constructed using Pecan St Dataport data \citep{noauthor_dataport_nodate} for residential appliances and data from two buildings in California. Device-level demand estimates can also be obtained with Non-Intrusive Load Monitoring (NILM) algorithms. The DFS model uses weighted predictions from a bank of models to separately predict AC and OL demand. These two predicted demands can be added and compared to the true feeder demand, allowing for real-time improvements to the weighted predictions.

\citet{marchesoni-acland_end--end_2020} construct an end-to-end NILM system using high frequency data and neural networks. They use the PLAID dataset \citep{medico_voltage_2020} to select the most important high-frequency features, and train artificial neural networks (ANNs) on the UK-DALE \citep{kelly_uk-dale_2015} dataset to disaggregate the household-level signal. They constructed a high-frequency meter to be attached at the household level for 2 houses in Uruguay and smart meters at the device level, sampling every 1 minute, for evaluation. This approach and the successful results seem the most relevant to our problem. The ‘rectangles’ network in the paper predicts the beginning and end of the appliance  activation and the power consumed. We instead predict the number of running medical devices of a fixed type rather than disaggregating the exact signal.

\citet{bucci_cnn_2021} focus on detecting and classifying change of state events for different appliances from the aggregate current signal. They decompose the derivative of the RMS current using the Short Term Fourier Transform, which allows identification of each event based on its spectral information.

\section{Challenges and limitations}
\label{appendix:challenges-limitations}
The model predicts the number of medical devices downstream of a feeder. If this has to be mapped to the number of medically fragile individuals, further work in determining the number of devices per person is necessary, which may depend on data such as underlying conditions, geographical location, and income level of the neighborhood. Additionally, some common non-medical devices can be medically critical for some individuals. For example, for a diabetic person, the refrigerator is a critical medical device since it is necessary to store insulin.

The model will be trained to predict the number of devices belonging to a set of common in-home medical devices. Therefore, if none of the devices an individual owns belong to this set, the individual will not be accounted for. This will limit the transferability of this model to a case where devices other than those considered need to be identified.

The presented approach aims at extending the load disaggregation framework for feeder-level data. Although feeder level disaggregation has been implemented, for instance to identify air conditioning loads \citep{ledva_real-time_2018},  it remains to be seen if this can be successfully implemented for medical devices, particularly if the power signatures of other devices are similar to the target medical devices.

\end{document}